\documentclass[conference]{IEEEtran}
\IEEEoverridecommandlockouts
\usepackage{cite}
\usepackage{amsmath,amssymb,amsfonts}
\usepackage{algorithmic}
\usepackage{graphicx}
\usepackage{textcomp}
\usepackage{xcolor}
\def\BibTeX{{\rm B\kern-.05em{\sc i\kern-.025em b}\kern-.08em
    T\kern-.1667em\lower.7ex\hbox{E}\kern-.125emX}}

\begin{document}
\title{Neural Radiance Fields (NeRFs): A Review and Some Recent Developments\\
}

\author{\IEEEauthorblockN{Mohamed Debbagh}
\IEEEauthorblockA{
\textit{McGill University}\\
Montreal, Canada \\
mohamed.debbagh@mail.mcgill.ca}

}

\maketitle
\thispagestyle{plain}
\pagestyle{plain}
\begin{abstract}
Neural Radiance Field (NeRF) is a framework that represents a 3D scene in the weights of a fully connected neural network, known as the Multi-Layer Perception(MLP). The method was introduced for the task of novel view synthesis and is able to achieve state-of-the-art photorealistic image renderings from a given continuous viewpoint. NeRFs have become a popular field of research as recent developments have been made that expand the performance and capabilities of the base framework. Recent developments include methods that require less images to train the model for view synthesis as well as methods that are able to generate views from unconstrained and dynamic scene representations. 
\end{abstract}


\section{Introduction}
One of the major problems explored in computer vision research is view synthesis, which has many implications and shared methods within the field of computer graphics and 3D rendering. Solutions to this problem aim to develop an approach that can generate novel views of a particular scene given an input of 2D RGB images from a sparse set of viewpoints. The output of such a model should be sampled over a continuous set of viewpoints resulting in realistic novel views of the same scene. Some popular classes of approaches include light field interpolation, surface estimation through mesh-based approximations, and more recently neural volume rendering (neural network based approach). Neural Radiance Fields (NeRF) were introduced by Mildenhall et al. and fits within the latter class of approaches that use a neural network architecture to represent a scene and synthesize novel views using neural volume renderings in order to achieve state-of-the-art results. The original paper \cite{mildenhall2020nerf} has popularized NeRF as the prominent method for view synthesis and makes 3 overall contributions that allow the framework to produce photorealistic outputs that can model the complex shapes and representations of a scene. (1) The first being the representation of a continuous scene though a  simple fully connected neural network that maps a 5D input (3 euclidean coordinate dimensions and 2 viewing direction dimensions) to a 4D output (RGB color channels and volume density). (2) The second contribution is the use of neural volume rendering techniques that utilizes differentiable camera rays, which make optimization of RGB representations possible. (3) Finally, the use of positional encoding techniques to transform the input domain into a higher dimensional space, which allows for  the neural network to capture higher frequency details in the scene representation during training. The NeRF model has since been improved and expanded upon to capture various modes of representations. This paper reviews the original NeRF framework, referred to as vanilla NeRF, and further explores a few of the many contributions that have been made to expand the baseline model. This review will include the following NeRF based developments: PixelNeRF, RegNeRF, Mip-NeRF, Raw NeRF, and NeRF in-the-Wild. For the sake of reviewing these concepts at the high level, we will not include the specific equations or model architectures that were designed for experimentation and we suggest that you explore the original papers for details on specific implementations.
\begin{figure*}[t]
\normalsize

\centerline{\includegraphics[scale=0.35]{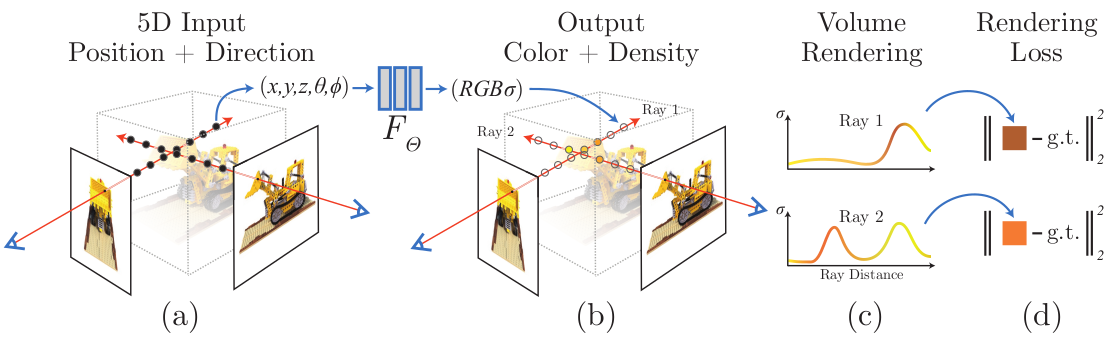}}
\caption{Overview of the NeRF scene representation pipeline. (a) Feed forward pass of the 5D input. (b) Mapping of 4D output in the 2D space. (c) Ray marching for volume rendering. (d) optimization over reconstruction loss. \cite{mildenhall2020nerf}}
\label{nerf}
\end{figure*}

\section{Neural Radiance Fields}
\subsection{Neural Volume Rendering}
The NeRF representation is built upon neural volumes, an implicit volumetric representation of a 3D scene learned and stored as the weights of a deep neural network. In Lombardi et al.\cite{Lombardi:2019} 2D images are fed into a Variational Autoencoder (VAE) and are encoded as a latent code. The output of the decoder reconstructs the latent code into a volumetric voxel representation with a RGB and alpha channel composite representation at each point in space. While the goal of their research was to construct a 3D representation from 2D images, the VAE is trained by reconstructing the 2D images from the voxel representation, using ray marching techniques for volume rendering. Ray marching is a differentiable process which makes optimization using gradient descent methods possible. A 2D reconstruction of a scene can be performed by estimating the radiance values of each pixel in the image plane that is projected from the 3D scene at a given viewing direction. 

Rays at each pixel site project into the 3D space at a given viewing direction perpendicular to the image/camera plane and are used to represent the volume or occupancy of the space along the ray. The volume density of the pixel is determined by taking the integral of volumes along the ray; this process is known as volume rendering. Since it is not computationally possible to determine volumes along a continuous ray, volumes along discrete points along the ray are sampled to estimate the integration; this technique for volume rendering is known as ray marching. The radiance and volume of the pixel are normally represented as color (r,g,b) and opacity in the reconstructed 2D image. This process can be mapped through neural networks, and the entirety of this representation method is known as neural volume rendering.

In the case of Lombardi et al. neural volume rendering  is used to reconstruct images from the 3D voxel output obtained from the output of the VAE.  However, due to the nature of VAE’s, artifacts and reconstruction warping occurs as the latent code from low dimensional space is being upsampled into the high dimensional 3D voxel space. Various additional techniques need to be applied to mitigate these effects. Thus the final 3D shape geometry is often left less than perfect. In contrast, the goal of view synthesis is to generate photorealistic images from novel viewpoints. Using VAEs to reconstruct a 3D voxel interpretation is unnecessary since it would lead to imperfections as described in \cite{Lombardi:2019}.

\subsection{NeRF 3D Scene Representation}
The original NeRF paper proposes that the representation of a scene be a neural volume that is described by the weights of a simple fully connected neural network architecture known as the Multilayer Perceptron (MLP), whose 5D inputs $(x,y,z,\theta, \Phi)$ correspond to a position in 3D space, $x=(x,y,z)$ and  2D viewing direction, $d=(\theta, \Phi$) that correspond to a point along a camera ray. The output of the MLP corresponds to a mapping of the color channel, $c=(r,g,b)$ and volume density, $\sigma$ of a pixel in the 2D image plane at that viewpoint. Unlike the previous study, the 3D representation of the scene is completely represented implicitly through the weights of the simple feed forward MLP and not through a voxel representation. The MLP feed forward network can be expressed as $F_{\Theta}:(x,d) \rightarrow (c,\sigma)$. The parameters, $\Theta$, of the MLP are optimized with a differentiable volume rendering function and  trained on a set of ground truth images and their known viewing directions. The loss function can be selected by evaluating differences between the true pixel color and expected pixel color from the volume rendering process. In the paper, the authors used a simple mean squared error. For a visual overview of the NeRF scene representation from the original paper \cite{mildenhall2020nerf} see Fig.\,\ref{nerf}.

\subsection{Positional Encoding}
When training the MLP, $F_{\Theta}:(x,d) \rightarrow (c,\sigma)$  directly as described in the previous section, the model tends to struggle to output highly detailed results. This is a common problem  in many encoding tasks, where one wants to encode a representation, such as an image, through the weights of  a fully connected neural network. This task is difficult as the MLP’s are biased  to learn low frequencies faster. Meaning, these networks tend to work faster on tasks that want to generalize outcomes, and avoid overfitting the data. However, since the goal of neural volume rendering is to fit a precise geometric shape to a 3D scene, it is better for the network to overfit the data. \cite{tancik2020fourfeat} Tancik et al. introduce a method, commonly used in transformers, called positional encoding, to map the low frequency input to a high frequency domain. Mapping the input to a high frequency domain allows the MLP to capture the high frequency and high resolution detail in the scene. When applied to the MLP, the NeRF model becomes $F_{\Phi}:(\gamma(x),\gamma(d)) \rightarrow (c,\sigma)$. Where $\gamma(.)$ is the function that maps our input into the high frequency domain. In the case of NeRF a Fourier feature mapping is used as the high frequency feature mapping function. Note that this is necessary to achieve the photorealistic results obtained from the NeRF model.

\begin{figure*}[t]
\normalsize

\centerline{\includegraphics[scale=0.19]{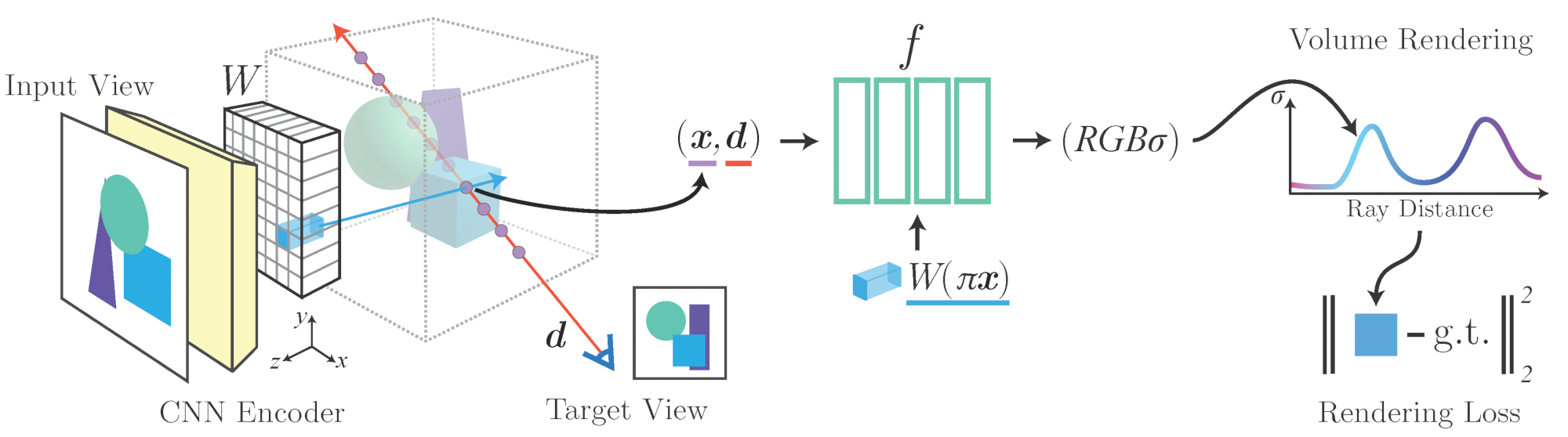}}
\caption{Overview of the PixelNeRF input and volume rendering pipeline. \cite{Yu_2021_CVPR}}
\label{pixelnerf}
\end{figure*}

\subsection{Properties}
Thus far, the NeRF model and its optimization method for view synthesis have been described as a neural volume representation that can capture high frequency geometric details of a 3D  scene. This gives NeRF some interesting intrinsic properties that extend beyond the task of view synthesis. The first attribute to note is that since the 3D geometric representation is stored as weights within a fully connected neural network, the NeRF can be considered a compressed format for 3D models. The 3D model can be reconstructed by querying the pre-trained NeRF over pre-defined viewpoints, then applying 3D geometric construction methods such as marching cubes. This is significant since the size of a NeRF file would be smaller than a single image that the model was trained on. The second attribute is that NeRF captures relational geometric information about a scene. This gives us the ability to use the detailed geometric information for tasks such as generating depth maps and shape visualizations. This can also be used to capture occlusion effects on mixed reality scenes. The third attribute is the ability to visualize the effect of color perception at different viewing directions. This attribute allows scenes to be captured under various lighting conditions in a photorealistic manner given a fixed position.

\subsection{Implementation and Challenges}
The resulting view points obtained from the NeRF approach are highly detailed and outperform prior state-of-the-art methods in both synthetically modeled scenes and real scenes. However the vanilla NeRF model described thus far has several constraints for real world implementation. The first aspect that we will focus on is the training and optimization process. The challenge with the implementation of the NeRF model is that each scene requires training on images with known viewpoint directions. While this problem seems quite limiting for on-the-go application, there exists methods to estimate these parameters including the COLMAP structure-from-motion package \cite{schoenberger2016sfm}. This may introduce some variation during the generation of new scenes, nevertheless the results obtained are still quite impressive. The training and rendering processes are very slow compared to other methods and require a diverse set of images from unique viewpoints to capture seamless continuous view synthesis. Most implementations require at least 80 images for training. Models trained with very sparse images will generate uninterpretable scenes and fail to generalize. Other challenges include the constraints on the scenes being captured. NeRFs are constrained to static scenes as the  influence of dynamic factors can have drastic effects on view synthesis. This includes reflections, moving objects and backgrounds. These challenges observed by the vanilla model have created a new field of study specifically on optimizing and expanding on the base NeRF concept. We will explore the recent developments that address some of these challenges in the next few section of this paper.

\section{View Synthesis From Fewer Images}
One challenge area that is addressed by recent developments in NeRF research is the calibration process of a scene. The implementation of NeRF is often limited due to the time and computational heavy process of training and rendering new scenes. One approach is to reduce the amount of resources spent on the calibration process. There are two papers that address the issue of reducing the amount of calibration images required.
\subsection{PixelNeRF}
The vanilla NeRF model requires many images from a diverse set of viewpoints as MLP model do not generalize well. MLPs also do not incorporate spatial information as images are flattened before they are fed into the training process. The vanilla method does not take into account information learned from multiple views if more than one viewpoint is used to calibrate the scene. This leads to the degradation of the scene synthesis when image sampling is not consistent and sparse (fewer than 80 images). Yu et al. \cite{Yu_2021_CVPR} introduces an expansion to the base NeRF model, which incorporates scene priors during calibration. This model is named PixelNeRF and the main contribution to the NeRF framework is the conditioning of the model on the input images by passing them through Convolutional Neural Networks (CNN) to train scene priors. To better illustrate this, a visual overview of the volume rendering pipeline from the paper ig given \cite{Yu_2021_CVPR}, see Fig.\,\ref{pixelnerf}. This allows the model to be trained with as low as one calibration image, though this is only recommended for simple geometries. In multi-view calibration (2 or more images) the output from the CNN of each input image at different views is combined before feeding it through the volume rendering process.  PixelNerf is able to achieve continuous scene representations on simple synthetic models from the ShapeNet database \cite{shapenet2015} with only one calibration image. The model was also tested on real images, and was able to generate a coherent geometric representation of the scene using a single calibration image, which is not possible from the vanilla NeRF. However, the results were not perfect and generated artifacts and distortions. This problem is significantly mitigated with the addition of multiple views (2-3 more) for calibration.

\subsection{RegNeRF}
Niemeyer et al. introduce an approach which reduces the floating artifacts and image inconsistencies that occur when the vanilla NeRF is trained only on a few images. The paper achieves this through the regularization of patches in unseen views for geometry smoothness and color \cite{Niemeyer2021Regnerf}. The model introduced by this paper is named RegNeRF and it improves upon the vanilla NeRF model optimization process. While the vanilla NeRF model optimizes over the reconstruction loss of the input images, it is not optimized to learn geometric consistencies at various points, thus the approach deteriorates as sample images become sparse. RegNeRF samples rays from patches at unseen viewpoints and then defines an optimization with the goal of regularizing the patches for geometry smoothness and color likelihood. This is done during the training process by defining loss functions of the regularization terms for both color and geometry patches. The results from this paper show significant improvement in reducing floating artifacts when compared to previous models. Since RegNerf keeps the MLP architecture of the original NeRF model, it is less computationally expensive during pre-training than the CNN based pixelNerf. RegNeRF can be trained using as low as 3 calibration images. 

\section{Dynamic and Unconstrained Conditions}
The dynamic conditions of a particular scene is a major contributing factor to its representation. Typically the vanilla NeRF model is not able to take advantage of these dynamic conditions, and in fact, it requires scenes to be constrained to achieve volumetric renderings without artifacts such a floating artifacts and aliasing. Recent developments to the NeRF model have explored methods for exploiting, controlling, and manipulating various aspects of a scene  conditions. In this section, we will explore a few papers that tackle issues in areas such as anti-aliasing at multi-scale representation, image processing pipe-line and representations from unconstrained sample images. 

\subsection{Mip-NeRF}
\begin{figure}[htbp]
\normalsize

\centerline{\includegraphics[scale=0.15]{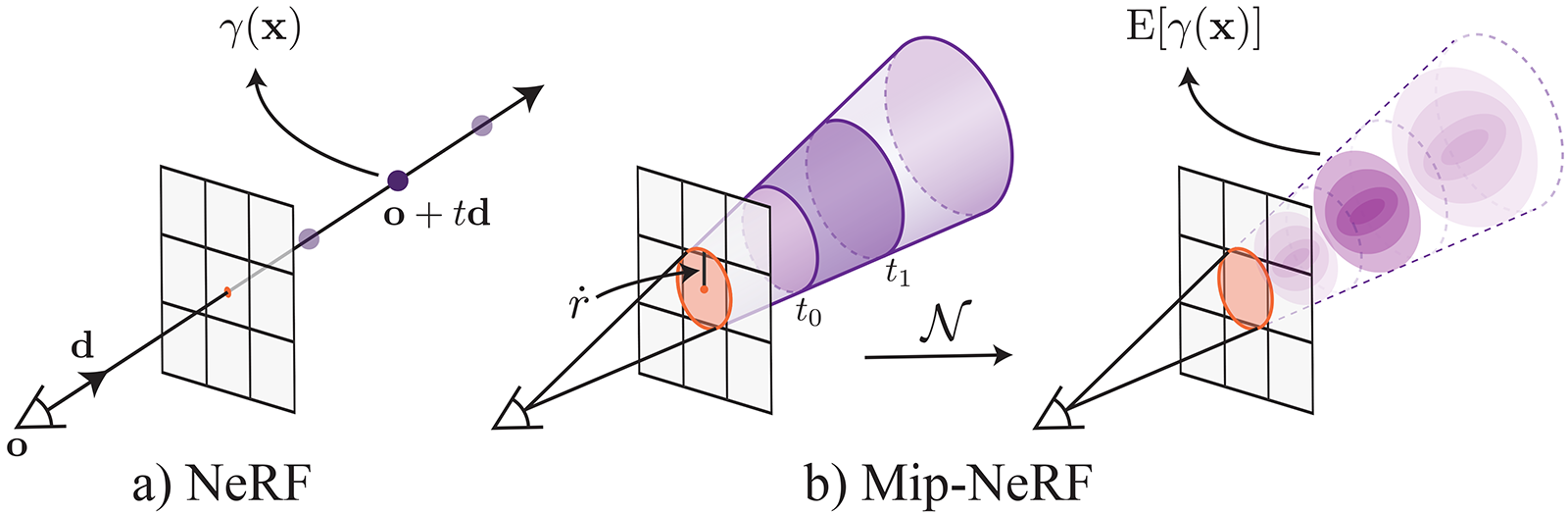}}
\caption{Comparison of NeRF ray marching and mip-NeRF ray cone Gaussian expectation sampling. \cite{barron2021mipnerf}}
\label{mipnerf}
\end{figure}

Multi-scale representation poses a challenge for many image processing and 3D rendering task. The recreation of a 3D scene or 2D image from different scales is often accompanied with artifacts, known as jaggies, which are often caused by aliasing. Aliasing is particularly observed in the NeRF model when sampling on lower resolution input images. The reconstruction of the views with the same resolution often contain these jaggies. Training NeRF models with multi-scale resolutions to mitigate this problem, often does not lead to significant improvements especially when trying to reproduce higher resolutions views. Barron et al. introduces Mip-NeRF, an extension of the NeRF approach that uses ray cones to capture a volume of space rather than a infinitesimal point to control multi-scale representation of a scene \cite{barron2021mipnerf}. As the scale of an image changes, so does the amount of information a single pixel captures from a scene. Thus, sampling points along a single point ray at each pixel causes distortions during interpolation with neighboring pixels, resulting in the aliasing effects. Sampling points along a regions in a conical ray allows for the capture of volumetric information in a non-linear way. The paper approximates these these conical intersection along these ray cones by fitting a multi-variate Gaussian distribution. Since sampling is no longer performed along a line, selecting samples in the distribution amounts to an expected value of the positional encoding, which in turn lets the network reason based on adjusted volume of space from scaling. A visual representation of the conical rays is given in a diagram from the original paper \cite {barron2021mipnerf}, see Fig.\,\ref{mipnerf}. The results from this study showed that Mip-NeRF outperforms multi-scale resolution reconstruction when compared to the previous vanilla NeRF method. It is also significantly more computationally efficient, when compared to the super sampling method with comparable results. 

\subsection{Raw NeRF}
\begin{figure}[htbp]
\normalsize

\centerline{\includegraphics[scale=0.4]{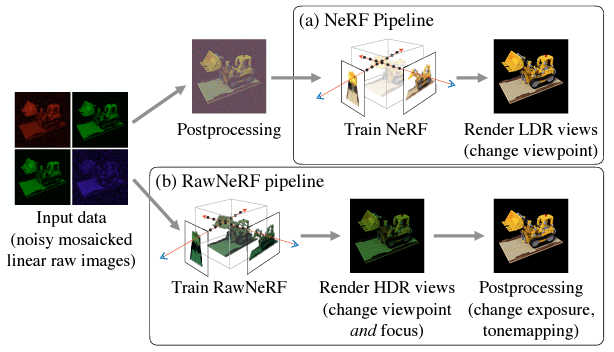}}
\caption{Overview of the rawNeRF input and volume rendering pipeline. \cite{mildenhall2022rawnerf}}
\label{rawnerf}
\end{figure}

In this section we take a look at a NeRF model approach that considers the image processing and post processing pipeline rather than the model architecture to gain more information from a scene resulting in impressive results. NeRF models are typically trained using Low Dynamic Range images (LDR) to perform novel view synthesis. This processing procedure is typically done to remove noise from images, especially in the dark. This is however at the expense of loss in details in darker regions of the scene. This loss of detail is reflected on novel views generated by the NeRF model. For example, scenes under very low lighting conditions, produce very dark viewpoint images with little to no detail. High Dynamic Range (HDR) images in contrast utilize a technique by combining multiple images with different exposures or views to capture details and even  apply post processing techniques for re-focusing. Mildenhall et al. proposes in their paper \cite{mildenhall2022rawnerf} that the inputs of the NeRF model be raw and minimally processed, noisy mosaicked linear images to capture more details of a scene, especially in the dark. The NeRF can then synthesis novel views points of the scene and apply post processing techniques to capture the effects similar to HDR in the final synthesized view. A visual representation of the Raw NeRF pipeline from the original paper is presented \cite{mildenhall2022rawnerf}, see Fig.\,\ref{rawnerf}. This methodology has many implications on novel view synthesis. First, the Raw NeRF is able to generate a denoised view of a scene that outperforms deep denoising methods used in LDR processing as will as multi-view denoising. Raw NeRF is able to render a scene with very low lighting conditions with photo-realistic detail. Moreover, post processing methods in the HDR color space can be applied to achieve further effects such as refactoring of exposure of a scene, tone mapping and refocusing. This is performed while also capturing the 3D geometric detail of a scene. 

\subsection{NeRF in the Wild}
A limitation of the vanilla NeRF model and  many of its variants is the requirements of constrained sampling conditions. This restricts many possible applications of NeRF on real world and natural images. This also constrains the NeRF model to fixed scenes with one or few objects and requires fairly consistent image viewpoints for calibration. When trained on unconstrained images and dynamic scenes, the NeRF generates views with floating artifacts, as it does know how to interpret these changing entities. These dynamic factors include photometric variations such as lighting conditions, and weather conditions as well as transient objects such as moving objects and temporary structures. Martin-Brualla et al. propose an expansion of the NeRF model called NeRF-W that embeds the static and transient components of a scene to generate novel views under dynamic conditions \cite{martinbrualla2020nerfw}. NeRF-W is able to disentangle the learned static components and dynamic factors by conditioning the inputs of the model on an appearance embedding and transient embedding. During training NeRF-W learns these interpretation by optimizing the embeddings along-side the weights of the NeRF on a reconstruction loss that is modulated with an uncertainty factor. Doing so NeRF-W is able to successfully isolate the structure of the scene with the dynamic aspects. Since the transient embedding is learned, scenes can be recreated under various conditions taking from the variety of the training data. In essence the NeRF-W is a disentangled version of the original NeRF model conditioned on the dynamic factors.

\section{Conclusion}
Since the development of the NeRF framework in 2020, many variants and expansions have been made that dramatically improve its performance and capabilities. The ability of the model to achieve state-of-the-art results and photo-realistic rendering presents many opportunities for such a framework in the field of view synthesis and beyond. NeRF has since become a field of research of its own with significant developments continuing to be made. Application of NeRFs include, 3D scene rendering in cinematography, 3D graphics generation, virtual rendering and walk-through of sites and many more. This paper covers a review of the base NeRF framework and explores some of the recent developments made thus far (at the time of writing this paper). It is highly recommended that each NeRF model variant be observed visually through video demonstration on the respective project sites as the capabilities can only be captured visually.

\bibliographystyle{ieeetran}
\bibliography{references}
\vspace{12pt}

\end{document}